\documentclass{article}

\usepackage{arxiv}
\usepackage{todonotes}
\usepackage[utf8]{inputenc} 
\usepackage[T1]{fontenc}    
\usepackage{hyperref}
\hypersetup{colorlinks,allcolors=black}      
\usepackage{url}            
\usepackage{booktabs}       
\usepackage{amsfonts}       
\usepackage{nicefrac}       
\usepackage{microtype}      
\usepackage{lipsum}
\usepackage{graphicx}
\usepackage{listings,xcolor}
\graphicspath{ {./images/} }
\usepackage{xcolor}
\usepackage{longtable}
\usepackage{caption}
\colorlet{punct}{red!60!black}
\definecolor{background}{HTML}{EEEEEE}
\definecolor{delim}{RGB}{20,105,176}
\colorlet{numb}{magenta!60!black}

\hypersetup{
    pdfauthor={Daniel Kershaw, Rob Koeling},
    pdftitle={ELSEVIER OA CC-BY CORPUS},
    pdflang={en-GB}
    }

\lstdefinelanguage{json}{
    basicstyle=\normalfont\ttfamily,
    numbers=left,
    numberstyle=\scriptsize,
    stepnumber=1,
    numbersep=8pt,
    showstringspaces=false,
    breaklines=true,
    frame=lines,
    backgroundcolor=\color{background},
    literate=
     *{0}{{{\color{numb}0}}}{1}
      {1}{{{\color{numb}1}}}{1}
      {2}{{{\color{numb}2}}}{1}
      {3}{{{\color{numb}3}}}{1}
      {4}{{{\color{numb}4}}}{1}
      {5}{{{\color{numb}5}}}{1}
      {6}{{{\color{numb}6}}}{1}
      {7}{{{\color{numb}7}}}{1}
      {8}{{{\color{numb}8}}}{1}
      {9}{{{\color{numb}9}}}{1}
      {:}{{{\color{punct}{:}}}}{1}
      {,}{{{\color{punct}{,}}}}{1}
      {\{}{{{\color{delim}{\{}}}}{1}
      {\}}{{{\color{delim}{\}}}}}{1}
      {[}{{{\color{delim}{[}}}}{1}
      {]}{{{\color{delim}{]}}}}{1},
}

\title{Elsevier OA CC-By Corpus}

\author{
 Daniel Kershaw\\
  Elsevier Ltd\\
  London, UK \\
  \texttt{d.kershaw@elsevier.com}\\
   \And
 Rob Koeling\\
  Elsevier Ltd\\
  London, UK \\
  \texttt{r.koeling@elsevier.com}\\
}

\begin{document}
\maketitle
\begin{abstract}
We introduce the Elsevier OA CC-BY corpus. This is the first open corpus of Scientific Research papers which has a representative sample from across scientific disciplines. This corpus not only includes the full text of the article, but also the metadata of the documents, along with the bibliographic information for each reference. 
\end{abstract}

\section{Introduction}
We are pleased to release the Elsevier OA CC-BY corpus for NLP and AI research. This is a corpus of $40k$ ($40,091$) open access (OA) CC-BY articles from across Elsevier’s journals representing a large scale, cross-discipline set of research data to support NLP and ML research. 

Research into the application of NLP and Machine Learning to scholarly content has attracted considerable attention in recent years. However, progress has been held back because of limited availability of large, cross-discipline datasets. Through releasing this dataset we hope to aid the research community in their work to expand the understanding of commonalities and differences between processing of scientific text and text of a different nature (e.g. news text). Moreover, this dataset allows research on challenges in processing scientific text that do not exist for other types of data. 

In this article we want to report on details of the dataset being released, focusing on the structure of the data, coverage of features and tools which can be used with it. 

\section{Related Work}
\label{sec:rel_work}

Open source datasets of academic articles are not new, notable corpora are Pub-Med OA corpus\footnote{\url{https://www.ncbi.nlm.nih.gov/pmc/tools/openftlist/}}, CiteSeerX~\cite{10.1145/276675.276685}, ACL~\cite{10.5555/1699750.1699759} and arXiv~\cite{saier2020unarxive}. The corpora mentioned above represent millions of academic articles freely available online in either PDF,~\LaTeX or free text. However, for these articles meta-data usually have not been resolved, meaning that establishing the connection between documents through citations is challenging. To overcome this, corpora such as the Literature Graph in Semantic Scholar have been released~\cite{Ammar_2018}, which has recently been superseded by S2ORC~\cite{lo-wang-2020-s2orc}. These corpora contain documents from multiple datasets (ACL, arXiv, PubMed $\cdots$) which have then been resolved by extracting references, venues and other metadata from the articles. However, the original datasets are biased towards certain academic disciplines and domains such as computing. The resulting literature graphs allow for the high resolution mapping of connections between entries found in academic articles, and therefore for the mapping of relations between authors, academic concepts and publication venues, to name a few. 

Similar to S2ORC are the datasets which have been compiled in response to world events such as COVID-19~\cite{Wang2020CORD19TC}. The COVID-19 dataset was derived from a number of already existing datasets and supplemented with additional published articles. As this dataset was designed for a particular cause, it is biased towards the biological and medial fields. The research output of the COVID19 dataset is extensive, including document summerisation~\cite{dan2020caire}, drug re-purposing~\cite{wang2020covid} and risk factor identification~\cite{wolinskiautomatic}. A comprehensive overview will take a review article in itself to summarise this contribution. This in essence shows the usefulness of domain specific corpora. 

The release of these types of datasets opened the doors for the development of task-specific datasets focused on curating data for one or more specific tasks. For example, \cite{Cohan2019Structural} released labelled data for developing models for citation intent classification. Whereas~\cite{jain-etal-2020-scirex} released annotated articles to develop document level information extraction systems. Specific task datasets for co-reference resolution and entity extraction usually come from shared tasks from workshops such as SemEval, TREC and BioNLP. High-quality annotations are  required for these tasks, which can be challenging (meaning time consuming and expensive) to produce. The corresponding datasets are generally small, highly curated and often focus on one particular phenomenon, topic domain or text genre. Very good overviews of available datasets for typical tasks are presented in \cite{GOYAL201821} (for Named Entity Recognition) and in \cite{SUKTHANKER2020139} (for co-reference resolution). These overviews show that datasets containing academic writing are mostly limited to the bio-medical domain. Moreover, the limited size of these data sets constitute a serious limitation. Even when models developed on the basis of this data show promising results, there are limited options to validate the results across domains or using larger corpora.

\section{Dataset}

This section details how the data set was constructed, which fields are included and coverage of fields in the dataset.

\subsection{Data Sampling}
The corpus consists of scientific articles published by Elsevier since the beginning of $2014$ which are open access (OA) and covered by a CC-BY 4.0 licence.\footnote{This license allows users to copy, to create extracts, abstracts and new works from the article, to alter and revise the article, and to make commercial use of it (including re-use and/or resale of the article by commercial entities), provided the user gives appropriate credit (with a link to the formal publication through the relevant DOI), provides a link to the license, indicates if changes were made and the licensor is not represented as endorsing the use made of the work. \url{https://www.elsevier.com/about/policies/open-access-licenses/user-licences}} In order to create a balanced dataset from across academic disciplines, a stratified sampling method was used to achieve equal representation from across Elsevier disciplines as represented by ASJC (All Science Journal Classification) codes. ASJC codes represent the scientific discipline of the journals in which the article was published. To simplify this, the 334 ASJC codes were grouped into their 27 top-level subject classifications.\footnote{Serial titles are classified using the ASJC scheme. This is done by in-house experts at the moment the serial title is set up for Scopus coverage; the classification is based on the aims and scope of the title, and on the content it publishes. \url{https://service.elsevier.com/app/answers/detail/a_id/15181/supporthub/scopus/}}

$2000$ documents were sampled from each of the 27 top-level ASJC classes. Each article can have multiple ASJC codes, if the article was selected for one class then it was removed from the remaining pool. The resulting sample of documents is balanced across disciplines (see Table~\ref{tab:asjc_dis} for the break down of document per discipline)

The base data-set from which articles were sampled was a cleansed corpus. This means that the articles had to have a minimum of 20 sentences, and the sentence had to be `clean', meaning that if the sentence has an excessive amount of \texttt{XML} or other markup language then it was removed from the article. A maximum of 20\% of sentence can be removed from an artical to be included in the base corpus.\footnote{The cleaning of sentences mainly effects documents from the Mathematics domain as these have large amounts of \LaTeX markup within sentences.}

\subsection{Data Structure}

Each document within the corpus is contained within its own JSON file. The name of the file is the ID of the article. The data for each article is structured as described in the JSON scheme and field descriptions below. 

\begin{lstlisting}[language=json,firstnumber=1]
{
    "docId": <str>,
    "metadata":{
        "title": <str>,
        "authors": [
            {
                "first": <str>,
                "initial": <str>,
                "last": <str>,
                "email": <str>
            },
            ...
        ],
        "issn": <str>,
        "volume": <str>,
        "firstpage": <str>,
        "lastpage": <str>,
        "pub_year": <int>,
        "doi": <str>,
        "pmid": <str>,
        "openaccess": "Full",
        "subjareas": [<str>],
        "keywords": [<str>],
        "asjc": [<int>],
    },
    "abstract":[
        {
          "sentence": <str>,
          "startOffset": <int>,
          "endOffset": <int>
        },
        ...
    ],
    "bib_entries":{
        <str>:{
            "title":<str>,
            "authors":[
                {
                "last":<str>,
                "initial":<str>,
                "first":<str>
                },
                ...
            ],
            "issn": <str>,
            "volume": <str>,
            "firstpage": <str>,
            "lastpage": <str>,
            "pub_year": <int>,
            "doi": <str>,
            "pmid": <str>
        },
        ...
    },
    "body_text":[
        {
        "sentence": <str>,
        "secId": <str>,
        "startOffset": <int>,
        "endOffset": <int>,
        "title": <str>,
        "refoffsets": {
            <str>:{
                "endOffset":<int>,
                "startOffset":<int>
                }
            },
        "parents": [
            {
            "id": <str>,
            "title": <str>
            },
            ...
        ]
    },
    ...
    ]
}
\end{lstlisting}

\paragraph{docId}
The docID is the identifier of the document. This is unique to the document, and can be resolved into a URL for the document through the addition of https//www.sciencedirect.com/science/pii/<docId>

\paragraph{abstract}
This is the author provided abstract for the document

\paragraph{body\_text}
The full text for the document. The text has been split on sentence boundaries, thus making it easier to use across research projects. Each sentence has the title (and ID) of the section which it is from, along with titles (and IDs) of the parent section. The highest-level section takes index 0 in the parents array. If the array is empty then the title of the section for the sentence is the highest level section title. This will allow for the reconstruction of the article structure. References have been extracted from the sentences. The IDs of the extracted reference and their respective offset within the sentence can be found in the ``refoffsets'' field. The complete list of references are can be found in the ``bib\_entry'' field along with the references' respective metadata. Some will be missing as we only keep `clean' sentences, 

\paragraph{bib\_entities}
All the references from within the document can be found in this section. If the meta data for the reference is available, it has been added against the key for the reference. Where possible information such as the document titles, authors, and relevant identifiers (DOI and PMID) are included. The keys for each reference can be found in the sentence where the reference is used with the start and end offset of where in the sentence that reference was used. 
	
\paragraph{metadata}
Meta data includes additional information about the article, such as list of authors, relevant IDs (DOI and PMID). Along with a number of classification schemes such as ASJC and Subject Classification.

\paragraph{Author\_highlights}
Author highlights were included in the corpus where the author(s) have provided them. The coverage is 61\% of all articles. The author highlights, consisting of 4 to 6 sentences, is provided by the author with the aim of summarising the core findings and results in the article.

\subsection{Data Coverage and Statistics}
\begin{table}[!htb]
\begin{minipage}{.5\linewidth}
    \centering

    \caption{JSON Field coverage}
    \label{tab:first_table}

    \medskip

\begin{tabular}{@{}lr@{}} 
\toprule
\textbf{Field}          & \multicolumn{1}{l}{\textbf{Number of Articles}} \\ \midrule
Abstract                & 99.25                                 \\
Body\_text              & 100.00                                \\
Author\_highlight       & 61.31                                 \\
Metadata                & 100.00                                \\
Metadata - issn         & 100.00                                \\
Metadata - firstpage    & 85.50                                 \\
Metadata - lastpage     & 85.34                                \\
Metadata - pub\_year    & 100.00                                \\
Metadata - doi          & 100.00                                \\
Metadata - openaccess   & 100.00                                \\
Metadata - subjectareas & 100.00                                \\
Metadata - keywords     & 100.00                                \\
Metadata - asjc         & 100.00                                \\
Bib\_entries            & 97.60                                 \\ \bottomrule
\end{tabular}
\end{minipage}\hfill
\begin{minipage}{.5\linewidth}
    \centering

    \caption{Distribution of Publication Years}

    \medskip
\centering
\begin{tabular}{@{}ll@{}}
\toprule
\multicolumn{1}{l}{\textbf{Publication Year}} & \multicolumn{1}{l}{\textbf{Number of Articles}} \\ \midrule
2014                                 & 3018                                    \\
2015                                 & 4438                                    \\
2016                                 & 5913                                    \\
2017                                 & 6419                                    \\
2018                                 & 8016                                    \\
2019                                 & 10135                                    \\
2020                                 & 2159                                    \\ \bottomrule
\end{tabular}
\end{minipage}
\end{table}

\begin{table}[h]
\centering
\caption{Distribution of Articles Per Mid Level ASJC Code. Each article can belong to multiple ASJC codes.}
\label{tab:asjc_dis}
\begin{tabular}{@{}lr@{}}
\toprule
\textbf{Discipline}                          & \textbf{Count} \\ \midrule
General                                      & 3847           \\
Agricultural and Biological Sciences         & 4840           \\
Arts and Humanities                          & 982            \\
Biochemistry, Genetics and Molecular Biology & 8356           \\
Business, Management and Accounting          & 937            \\
Chemical Engineering                         & 1878           \\
Chemistry                                    & 2490           \\
Computer Science                             & 2039           \\
Decision Sciences                            & 406            \\
Earth and Planetary Sciences                 & 2393           \\
Economics, Econometrics and Finance          & 976            \\
Energy                                       & 2730           \\
Engineering                                  & 4778           \\
Environmental Science                        & 6049           \\
Immunology and Microbiology                  & 3211           \\
Materials Science                            & 3477           \\
Mathematics                                  & 538            \\
Medicine                                     & 7273           \\
Neuroscience                                 & 3669           \\
Nursing                                      & 308            \\
Pharmacology, Toxicology and Pharmaceutics   & 2405           \\
Physics and Astronomy                        & 2404           \\
Psychology                                   & 1760           \\
Social Sciences                              & 3540           \\
Veterinary                                   & 991            \\
Dentistry                                    & 40             \\
Health Professions                           & 821            \\ \bottomrule
\end{tabular}
\end{table}

\section{Use of Dataset}
The dataset~\cite{https://10.17632/zm33cdndxs.2} can be downloaded from Mendeley Data.\footnote{\url{https://data.mendeley.com/datasets/zm33cdndxs/3}} The dataset contains a formatted \texttt{JSON} version of the raw \texttt{XML} which can be accessed through the Elsevier APIs.\footnote{\url{https://dev.elsevier.com/}} The original \texttt{XML} files can be processed with the AnnotationQuery\footnote{\url{https://github.com/elsevierlabs-os/AnnotationQuery}} framework released by Elsevier Labs. 

Please use the following reference when using the dataset:
\begin{verbatim}
@dataset{https://10.17632/zm33cdndxs.3,
  doi = {10.17632/zm33cdndxs.2},
  url = {https://data.mendeley.com/datasets/zm33cdndxs/3},https://www.overleaf.com/project/5ef1aeeb7ff458000177cb45
  author = "Daniel Kershaw and Rob Koeling",
  keywords = {Science, Natural Language Processing, Machine Learning, Open Dataset},
  title = {Elsevier OA CC-BY Corpus},
  publisher = {Mendeley},
  year = {2020},
  month = aug
}
\end{verbatim}

\section{Conclusion and Future Work}
This dataset was released to support the development of ML and NLP models targeting science articles from across all research domains. While the release builds on other datasets designed for specific domains and tasks, it will allow for similar datasets to be derived or for the development of models  which can be applied and tested across domains. 

\section{Acknowledgements}
We would like to thank all our colleges at Elsevier for helping to pull together this dataset. Especially Darin McBeath from Elsevier Labs, and Anita Dewaard and Noelle Gracy from the Research Collaboration Office.

\bibliographystyle{unsrt}  
\bibliography{references}  

\end{document}